\documentclass[twocolumn, switch]{article} 

\usepackage{preprint}

\usepackage{amsmath, amsthm, amssymb, amsfonts}

\usepackage[numbers,square]{natbib}
\usepackage[utf8]{inputenc}	
\usepackage[T1]{fontenc}	
\usepackage{xcolor}		
\usepackage[colorlinks = true,
            linkcolor = purple,
            urlcolor  = blue,
            citecolor = cyan,
            anchorcolor = black]{hyperref}	
\usepackage{booktabs} 		
\usepackage{nicefrac}		
\usepackage{microtype}		
\usepackage{lineno}		
\usepackage{float}			
\usepackage{subcaption}

\usepackage{lipsum}		

\usepackage{newfloat}
\DeclareFloatingEnvironment[name={Supplementary Figure}]{suppfigure}
\usepackage{sidecap}
\sidecaptionvpos{figure}{c}

\usepackage{titlesec}
\titlespacing\section{0pt}{12pt plus 3pt minus 3pt}{1pt plus 1pt minus 1pt}
\titlespacing\subsection{0pt}{10pt plus 3pt minus 3pt}{1pt plus 1pt minus 1pt}
\titlespacing\subsubsection{0pt}{8pt plus 3pt minus 3pt}{1pt plus 1pt minus 1pt}

\usepackage{tikz,xcolor,hyperref}

\definecolor{lime}{HTML}{A6CE39}
\DeclareRobustCommand{\orcidicon}{
	\begin{tikzpicture}
	\draw[lime, fill=lime] (0,0) 
	circle [radius=0.16] 
	node[white] {{\fontfamily{qag}\selectfont \tiny ID}};
	\draw[white, fill=white] (-0.0625,0.095) 
	circle [radius=0.007];
	\end{tikzpicture}
	\hspace{-2mm}
}
\foreach \x in {A, ..., Z}{\expandafter\xdef\csname orcid\x\endcsname{\noexpand\href{https://orcid.org/\csname orcidauthor\x\endcsname}
			{\noexpand\orcidicon}}
}

\title{Development of a Realistic Crowd Simulation Environment for Fine-grained Validation of People Tracking Methods}

\usepackage{xwatermark}
\newwatermark[firstpage,color=gray!60,angle=90,scale=0.32, xpos=3.9in,ypos=0]{\href{https://doi.org/10.5220/0011691500003417}{\color{gray}{Publication doi: 10.5220/0011691500003417}}}
\newwatermark[firstpage,color=gray!90,angle=0,scale=0.28, xpos=0in,ypos=-5in]{*correspondence: \texttt{mstaniszewski@polsl.pl}}

\usepackage{authblk}

\author[1]{Paweł Foszner\orcidA{}}
\author[1]{Agnieszka Szczęsna\orcidC{}}
\author[2]{Luca Ciampi\orcidD{}}
\author[2]{Nicola Messina\orcidD{}}
\author[3]{Adam Cygan}
\author[3]{Bartosz Bizoń}
\author[4]{Michał Cogiel\orcidE{}}
\author[4]{Dominik Golba\orcidF{}}
\author[5]{Elżbieta Macioszek\orcidG{}}
\author[1\thanks{}]{Michał Staniszewski\orcidH{}}

\affil[1]{Department of Computer Graphics, Vision and Digital Systems, Faculty of Automatic Control, Electronics and Computer Science, Silesian University of Technology, Gliwice, Poland; name.surname@polsl.pl}
\affil[2]{Institute of Information Science and Technologies, National Research Council, Pisa, Italy; name.surname@isti.cnr.it}
\affil[3]{QSystems.pro sp. z o.o. Mochnackiego 34, 41-907 Bytom, Poland; nsurname@qsystems.pro}
\affil[4]{Blees sp. z o.o. Zygmunta Starego 24a/10, 44-100 Gliwice, Poland; nsurname@blees.co}
\affil[5]{Department of Transport Systems, Traffic Engineering and Logistics, Faculty of Transport and Aviation Engineering, Silesian University of Technology, Katowice, Poland; name.surname@polsl.pl}

\begin{document}

\twocolumn[ 
  \begin{@twocolumnfalse} 
  
\maketitle

\begin{abstract}
Generally, crowd datasets can be collected or generated from real or synthetic sources. Real data is generated by using infrastructure-based sensors (such as static cameras or other sensors). The use of simulation tools can significantly reduce the time required to generate scenario-specific crowd datasets, facilitate data-driven research, and next build functional machine learning models. The main goal of this work was to develop an extension of crowd simulation (named CrowdSim2) and prove its usability in the application of people-tracking algorithms. The simulator is developed using the very popular Unity 3D engine with particular emphasis on the aspects of realism in the environment, weather conditions, traffic, and the movement and models of individual agents. Finally, three methods of tracking were used to validate generated dataset: IOU-Tracker, Deep-Sort, and Deep-TAMA. 
\end{abstract}
\vspace{0.35cm}

  \end{@twocolumnfalse} 
] 

\section{Introduction}
\label{sec:introduction}

Using real crowd datasets can produce effective and reliable learning models, useful in the following applications such as object tracking \cite{Cafarelli_2022} \cite{focal_loss}, image segmentation \cite{Bolya_2019} \cite{deep_lab}, visual object counting \cite{counting_edge} \cite{video_counting} \cite{counting_cells}, individuals activity or violence recognition \cite{s22218345,pawel_foszner_2022_7044203}, crowd anomaly detection and prediction and wider crowd management solutions monitor. 
However, acquiring real crowd data faces several challenges, including the expensive installation of a sensory infrastructure, the data pre-processing costs, and the lack of real datasets that cover particular crowd scenarios. Consequently, simulation tools have been adopted for generating synthetic datasets to overcome the challenges associated with their real counterparts. Using simulation tools that can significantly reduce the time required to generate scenario-specific crowd datasets, mimic observed crowds in a realistic environment, facilitate data-driven research, and build functional machine learning models \cite{khadka2019learning,viped_2} based on generated data. Simulation offers flexibility in adjusting the scenarios, and generating and reproducing datasets with defined requirements. 

\begin{figure*}[!h]
 \centering
  \includegraphics[width=0.32\textwidth]{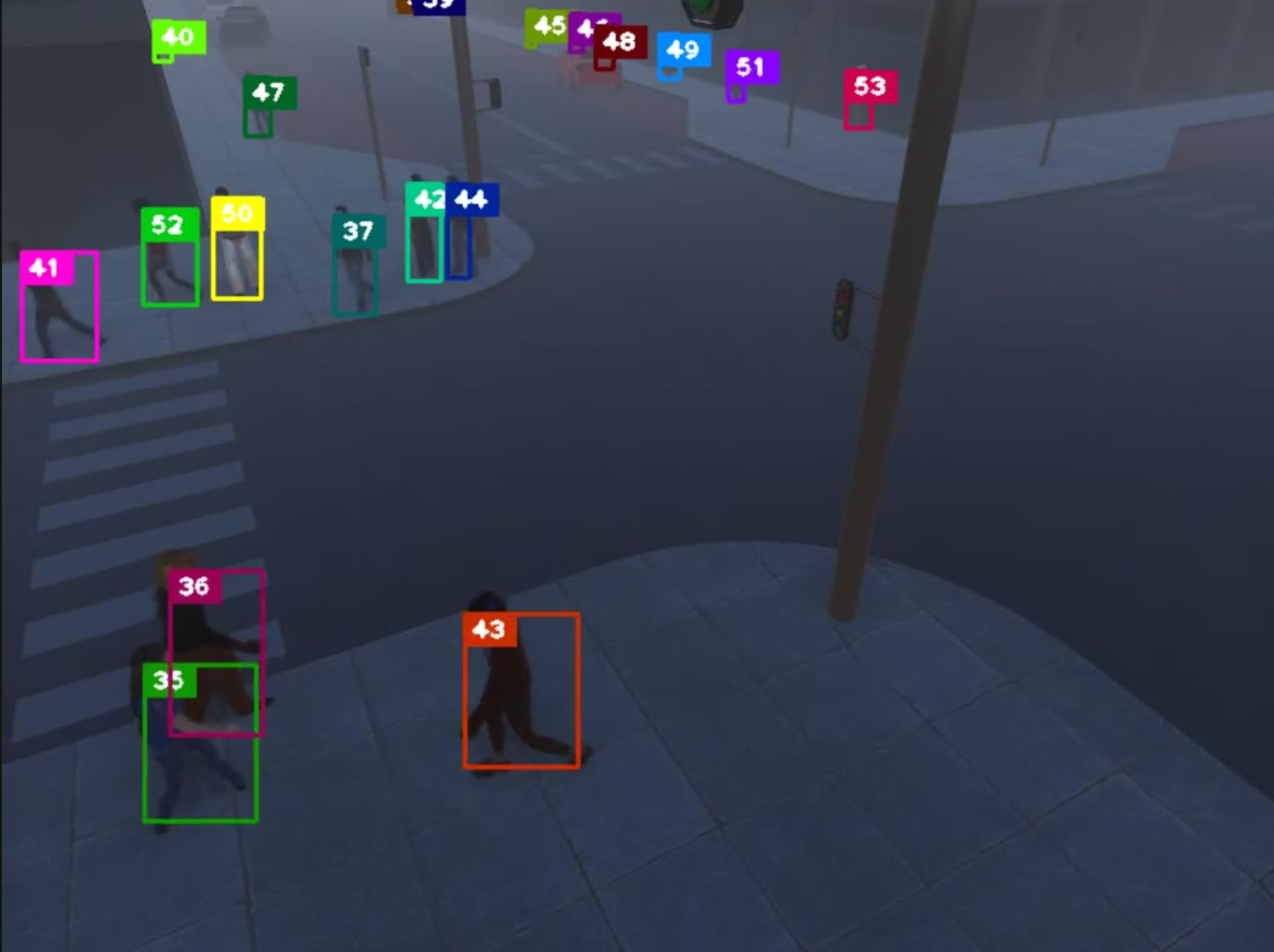}
  \includegraphics[width=0.32\textwidth]{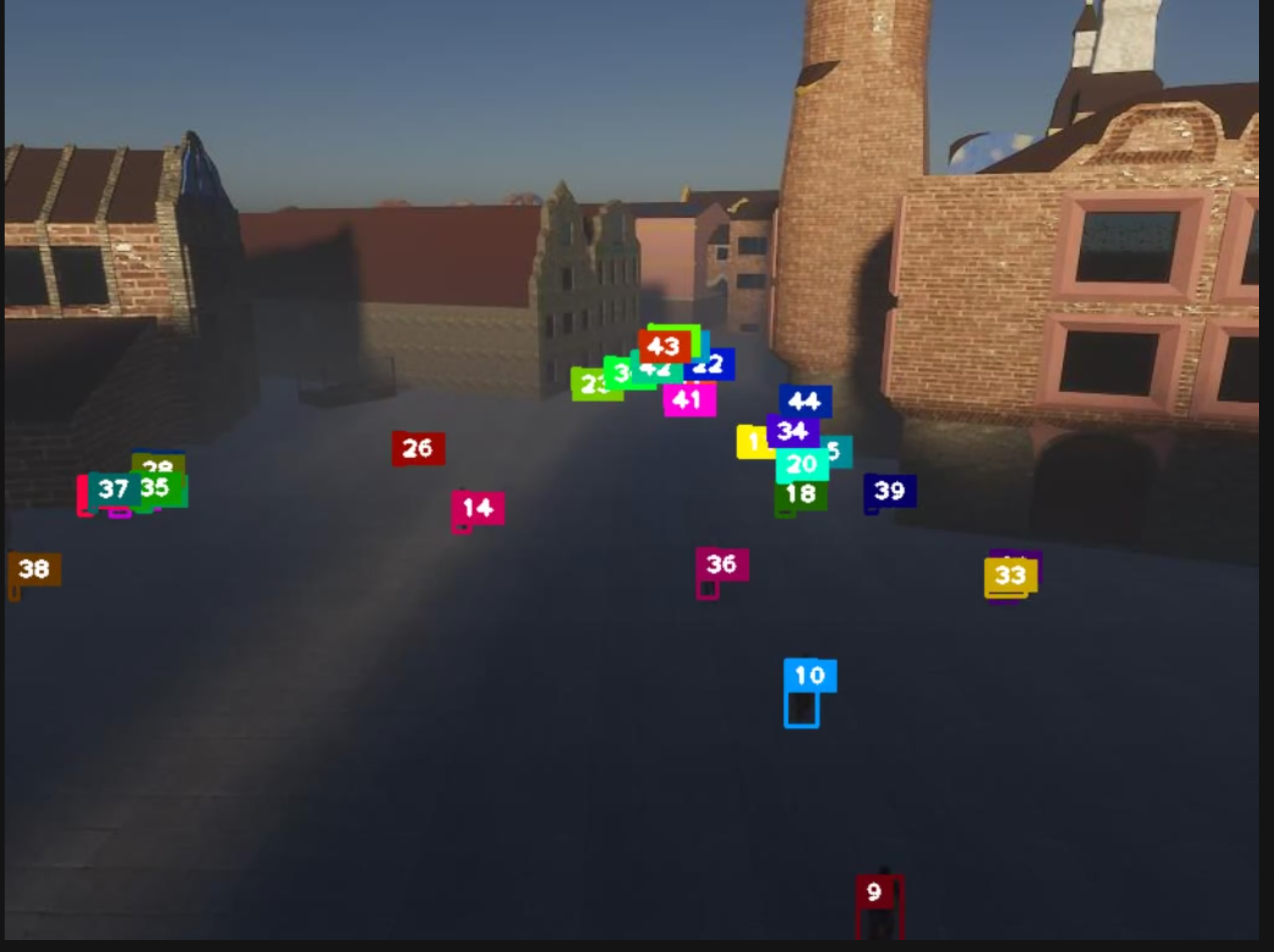}
  \includegraphics[width=0.32\textwidth]{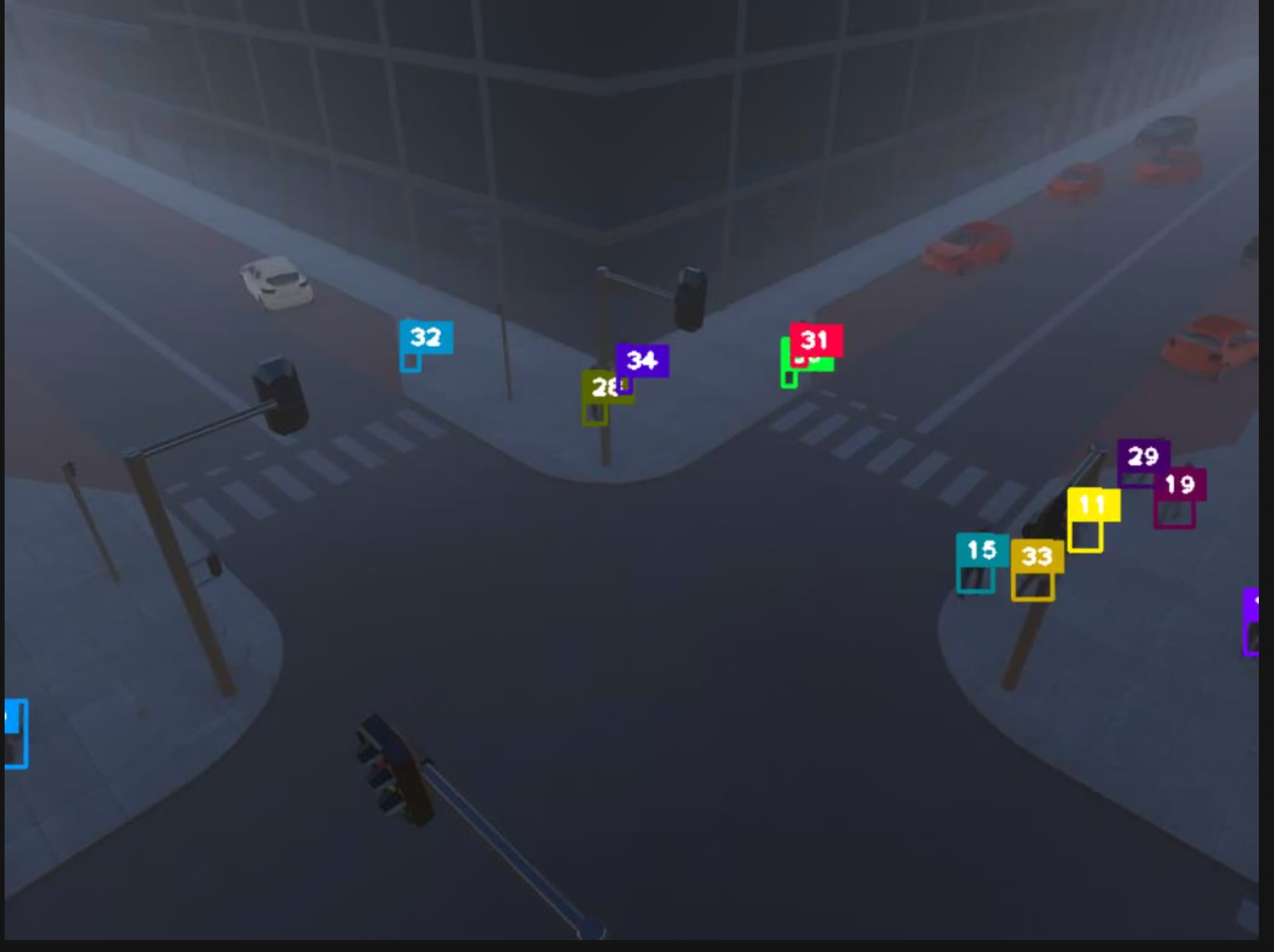}
 \caption{The visualization of tracking of multiple pedestrians in images generated by application of CrowdSim2.}
 \label{fig:tracking}
\end{figure*}




The main motivation for that work was to implement more realistic crowd simulation with additional features that can be applied in many modern artificial intelligence approaches (including the evaluation of people tracking algorithms). The proposed crowd simulator has the following advantages:
\begin{itemize}
    \item realism enhancement by application of motion matching, people and car generation,
    \item automatic assessment of ground truth and detection (Figure \ref{fig:tracking}) in the known format of MOT Challenge \cite{MOT20},
    \item simulation placed in 3 different localisations (with 3 views) for people movement and 2 localisations for cars,
    \item introduction of 4 realistic weather conditions including sun, fog, rain, snow, and different day time,
    \item many possible options for application including object detection and tracking, action detection, and recognition.
\end{itemize}

\begin{figure*}[!h]
 \centering
  \includegraphics[width=0.45\textwidth]{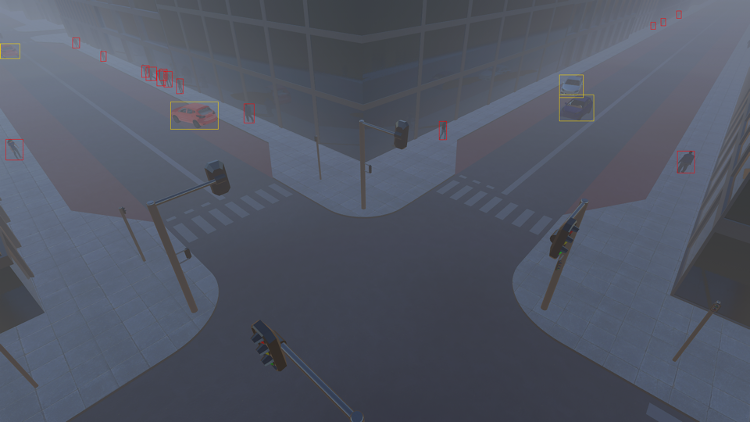}
  \includegraphics[width=0.45\textwidth]{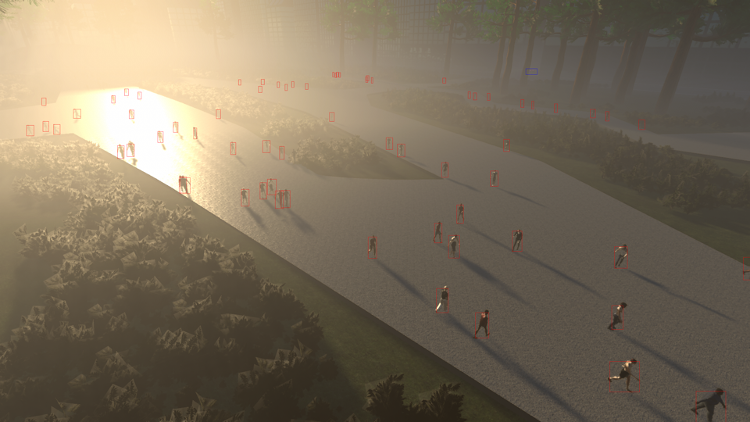}
  \includegraphics[width=0.45\textwidth]{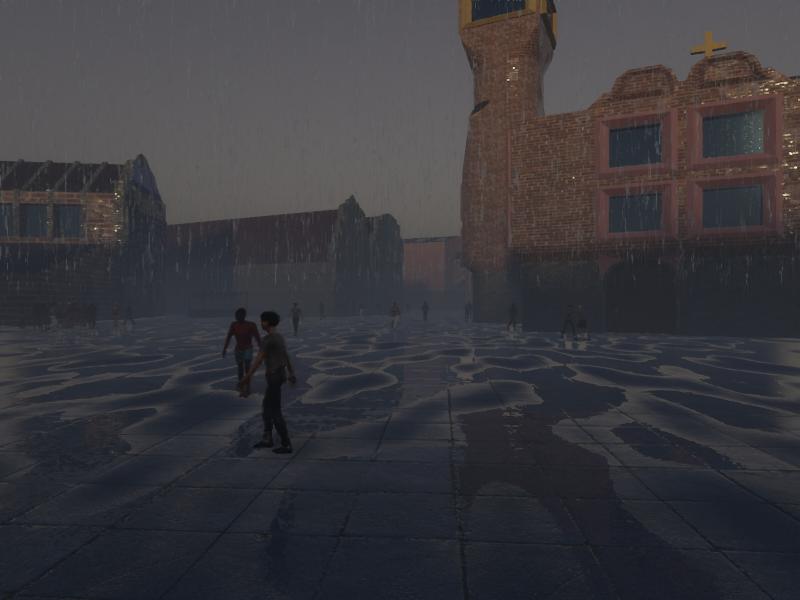}
  \includegraphics[width=0.45\textwidth]{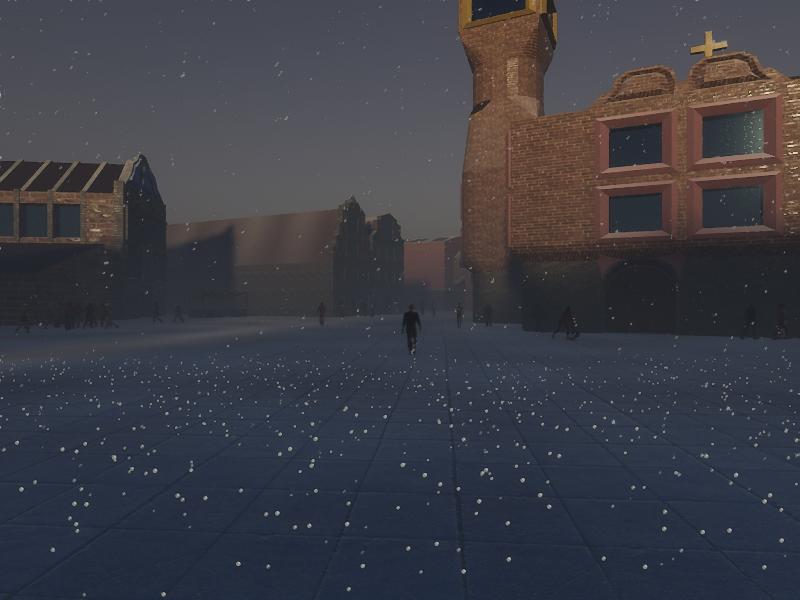}
 \caption{Exemplary views from CrowdSim2: junction and park with moving pedestrians and cars along with examples of snow and rain weather conditions that can be generated.}
 \label{fig:views}
\end{figure*}

\section{Related works}
Databases containing visual data are needed to develop the detection and tracking method, including cars and pedestrians.
For that purpose the annotations of traced objects (like a pedestrian) which includes an approximate bounding box are necessary.
The Mall dataset \cite{chen2012feature} was collected from a publicly accessible webcam with ground truth consisting of annotating 60,000 pedestrians. NWPU \cite{wang2020nwpu} includes approximately 5,000 images and 2,133,375 annotated heads. JHU-CROWD++ \cite{sindagi2020jhu} is another crowd dataset captured in different scenarios, and geographical locations, under weather conditions such as fog, haze, snow, and rain. JHU-CROWD++ provides head-level labeling which includes an approximate bounding box. The GTA5 Crowd Counting (GCC) \cite{wang2019learning} is an example of a large-scale visual synthetic dataset (15,212 images, 7,625,843 persons) generated using the well-known video game GTA5 (Grand Theft Auto 5). AGORASET \cite{courty2014using} is also a visual synthetic dataset for crowd video analysis. For a comprehensive overview of databases and simulators, see the following review articles
\cite{bamaqa2022simcd,lemonari2022authoring,van2021algorithms,yang2020review}. In \cite{amirian2020opentraj} the analysis of the statistical properties of real word datasets is available. 
Recent advancements in crowd simulation unravel a wide range of functionalities for virtual agents, delivering highly-realistic, natural virtual crowds. 

In the following work, simulated data will be used for the evaluation of different tracking algorithms \cite{tracking}. Here the concept of tracking by detection will be used, which means that detection will be available along with simulated data, and afterward tracking algorithms are applied for joining detections in tracks. Additionally, instead of whole-person detection, facial recognition could be applied \cite{peszor}. The first considered algorithm IOU-tracker was presented in \cite{1547Bochinski2018} and it doesn't use any image information, which allows it to run simpler tracking algorithms. Thanks to this non-image approach, it uses much less computing power than other trackers. The authors of the Deep-Sort method \cite{Wojke2018deep} - presented Simple Online and Realtime Tracking with a Deep Association Metric as a tracking-by-detection method. Deep-Sort is an extension of the SORT algorithm \cite{Wojke2017simple} made to integrate appearance information based on a deep appearance descriptor. The Deep-TAMA method \cite{DBLP:journals/corr/abs-1907-00831}– which stands for Deep Temporal Appearance Matching Association contrary to Deep-Sort and IOU-tracker during one stage performs tracking together with evaluation of the results. Another group of SORT applications - Observation-Centric SORT \cite{Bewley2016_sort} - is used for a multiple object tracker. OC-SORT was built to fix limitations in the Kalman filter and SORT algorithm. It is an online tracker and it has improved non-linear motion and robustness over occlusion. For wide application, the framework MMtracking \cite{mmtrack2020} was established which is an open-source video perception toolbox by PyTorch.

\section{Crowd simulator}

\begin{table}
\centering
\caption{Information summarizing a number of folders, seconds, and frames of data for different weather conditions.}
\label{tab:nums}
{%
\begin{tabular}{|c|c|c|c|}
\hline
Number of     & folders & seconds & frames \\ \hline
Sun  & 2899                 & 86 970                          & 2 174 250       \\ \hline
Rain & 1633                 & 48 990                          & 1 224 750         \\ \hline
Fog  & 1653                 & 49 590                          & 1 239 750       \\ \hline
Snow & 1646                 & 49 380                          & 1 234 500          \\ \hline
\end{tabular}
}
\end{table}

The proposed crowd simulator \textbf{CrowdSim2}\footnote{The dataset is freely available in the Zenodo Repository at \url{https://doi.org/10.5281/zenodo.7262220}} is the next version of crowd simulator \textbf{CrowdSim} \cite{staniszewski2020application} especially for testing multi-object tracking algorithms but also for action and object detection. It uses the microscopic (or ‘agent-based’) crowd simulation methods that model the behavior of each person, from which collective behavior can then emerge \cite{saeed2022simulating,van2021algorithms}. The simulator is developed using the very popular Unity 3D engine with particular emphasis on the aspects of realism in the environment, weather conditions, traffic, and the movement and models of individual agents. The proposed system can be used to generate a sequence of random images (datasets) for use in tracking and object detection algorithms evaluation but also in the crowd, car counting, and other crowd and traffic analysis tasks. The generated output data is in the format of the MOT challenge. The most important components of \textbf{CrowdSim2}, to support realism when rendering the resulting image, are described below.

\subsection{Agents Motion and Interactions}
\begin{figure}[!h]
 \centering
  \includegraphics[width=0.4\textwidth]{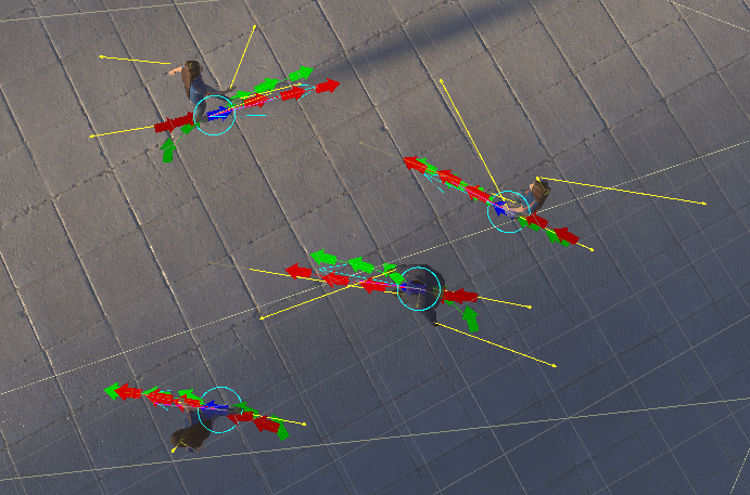}
 \caption{Agents movement concerning the concept of motion matching and system of features.}
 \label{fig:vector_person}
\end{figure}

\begin{figure}[!h]
 \centering
  \includegraphics[width=0.4\textwidth]{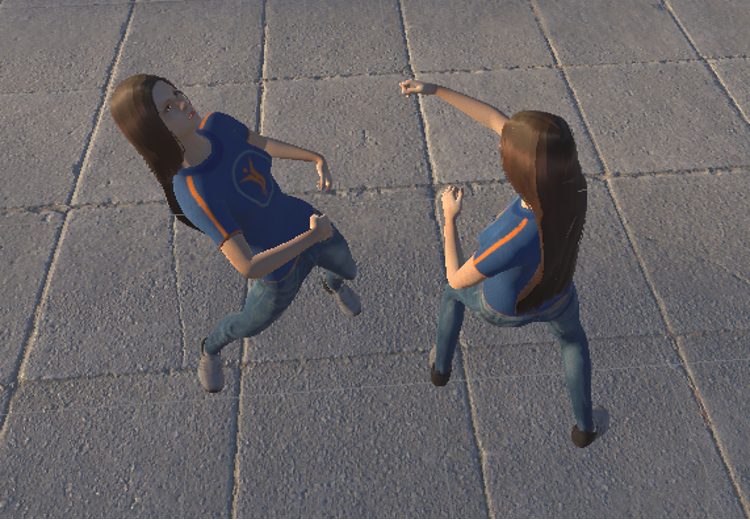}
 \caption{Random animated interaction between agents generated in CrowdSim2.}
 \label{fig:fight}
\end{figure}

The component necessary for producing lively and realistic virtual crowds is animating the characters, thus creating 3D moving agents. Data-driven approaches include methods utilizing motion capture data to use during skeleton-based animation of 3D human models \cite{elsa}. This approach requires many variations of data to represent movements in different activities. To ensure the universality of the system and to generate animations based on real human motion data, a motion matching algorithm was used \cite{clavet2016motion}. Motion matching is an alternative animation system without the need for a state machine with vectors given in Figure \ref{fig:vector_person}. Thanks to this, it is possible to perform different activities at the agent level, including dancing or fights. In the future, it is planned to use the learned motion matching algorithm \cite{holden2020learned} with additional styling \cite{aberman2020unpaired,holden2017fast}. Currently, we only have two styles of movement male and female, which rely on separate motion databases. 
The interactions are carried out based on interaction zones. These zones are placed in the city and define the type of interaction (for example dance, fight). When an agent enters the zone, a set of conditions is checked, and, depending on the situation, the agent is either ignored interaction, added to the queue, or starts interaction (see Figure \ref{fig:fight}). If an agent is added to the queue, there are not enough agents in the zone to start interacting. The agent moves normally while waiting in the queue, and if he moves too far away from the zone before the interaction begins, he is removed from the zone queue.

\subsection{Photo-realistic Rendering and Traffic}

\begin{figure}[!h]
 \centering
  \includegraphics[width=0.4\textwidth]{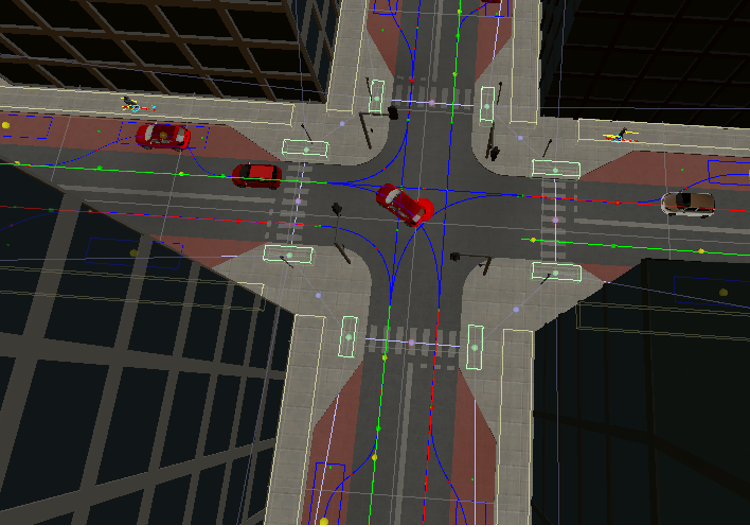}
 \caption{The navigation system for car movement with parking area and pedestrian stops.}
 \label{fig:vector_car}
\end{figure}

The main element is physical volumetric light that responds adequately to dynamically changing surroundings. There is a dynamic volumetric fog in the simulation. Additionally, the simulation also includes snow and rain based on particle effects (Figure \ref{fig:views}).
Finally, thanks to the use of the High Definition Render Pipeline (HDRP) in Unity engine and physical cameras,  it is possible to map the lens and matrix settings of the real camera to create photo-realistic output images. For the global agent movement on a macro scale, the already built-in NavMesh was used. On the micro-scale at the beginning of the study of human behavior, Emergent human behavior in Navigation was created by the application of the unique system of features. Cars can park in randomly selected parking bays as visible in Figure \ref{fig:vector_car}. At the crossing, they also choose a random direction. Traffic is based on a created system of nodes located on the roads, particularly crossing, sharp turns, and parking places. 




\begin{figure*}[!h]
 \centering
 \begin{subfigure}[b]{0.48\textwidth}
  \includegraphics[width=\textwidth]{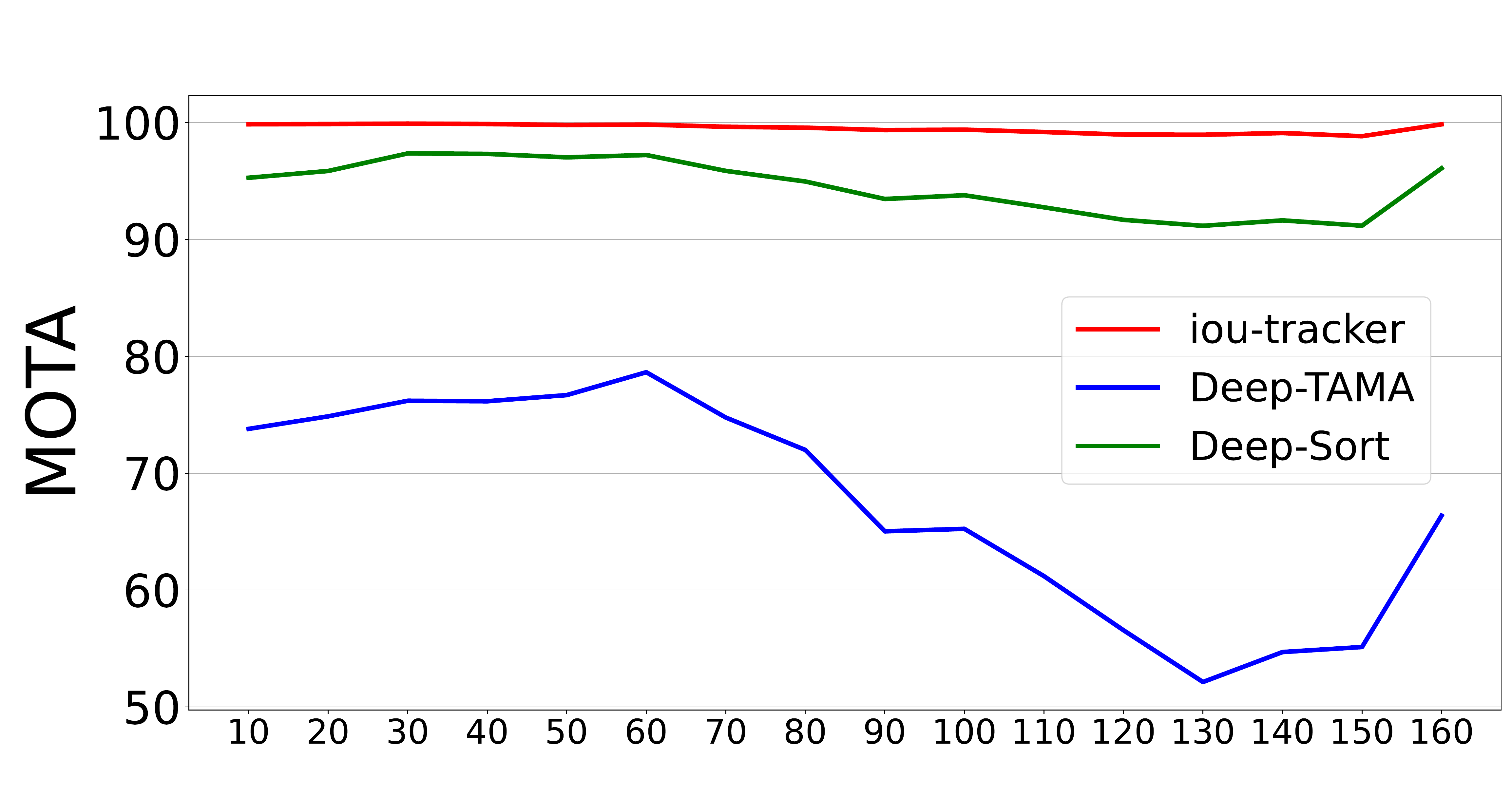}
  \caption{MOTA for varying crowd density.}
 \end{subfigure}
  \begin{subfigure}[b]{0.48\textwidth}
  \includegraphics[width=\textwidth]{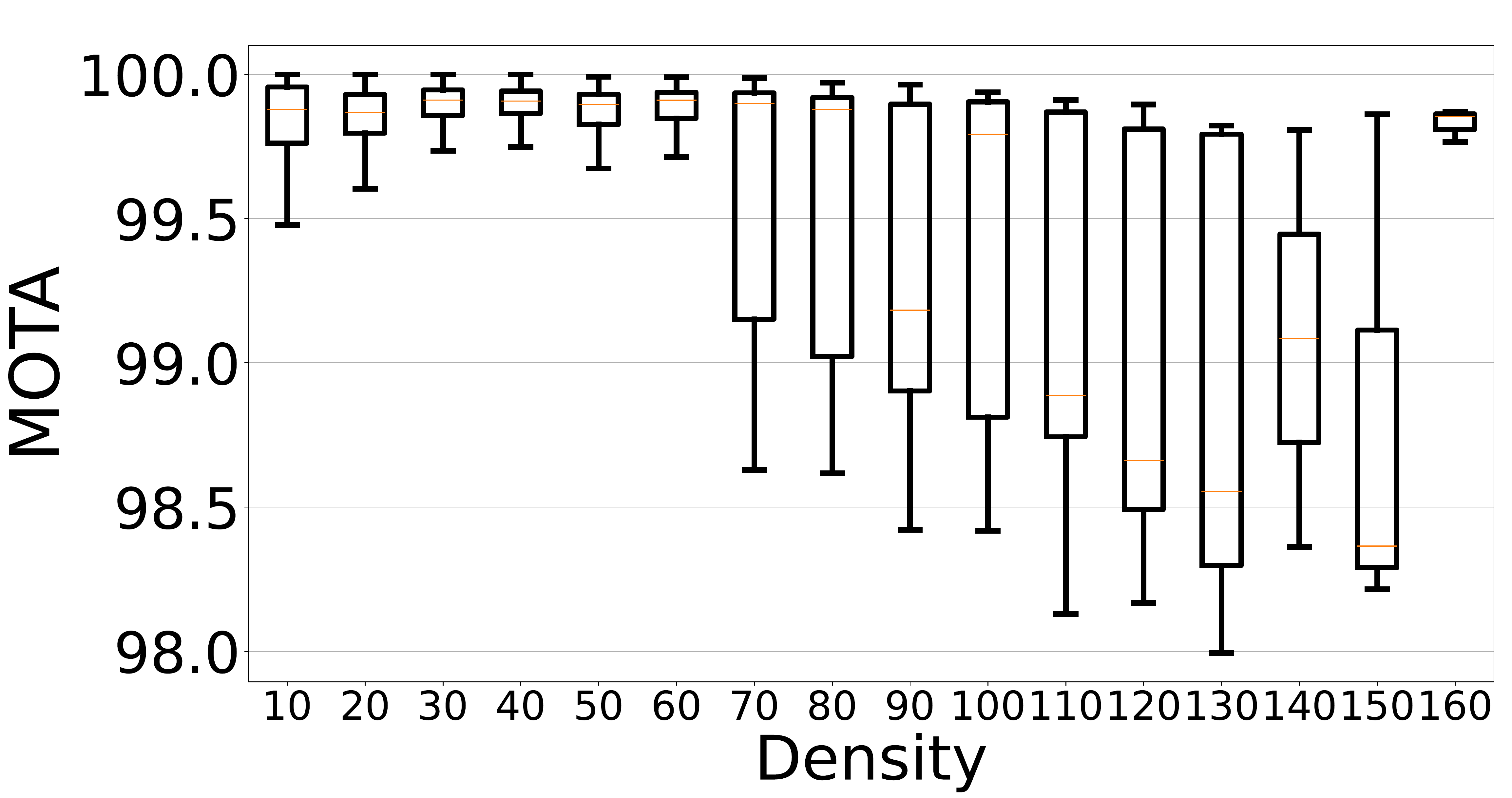}
  \caption{Varying density for MOTA on IOU-tracker.}
 \end{subfigure}
 \begin{subfigure}[b]{0.48\textwidth}
  \includegraphics[width=\textwidth]{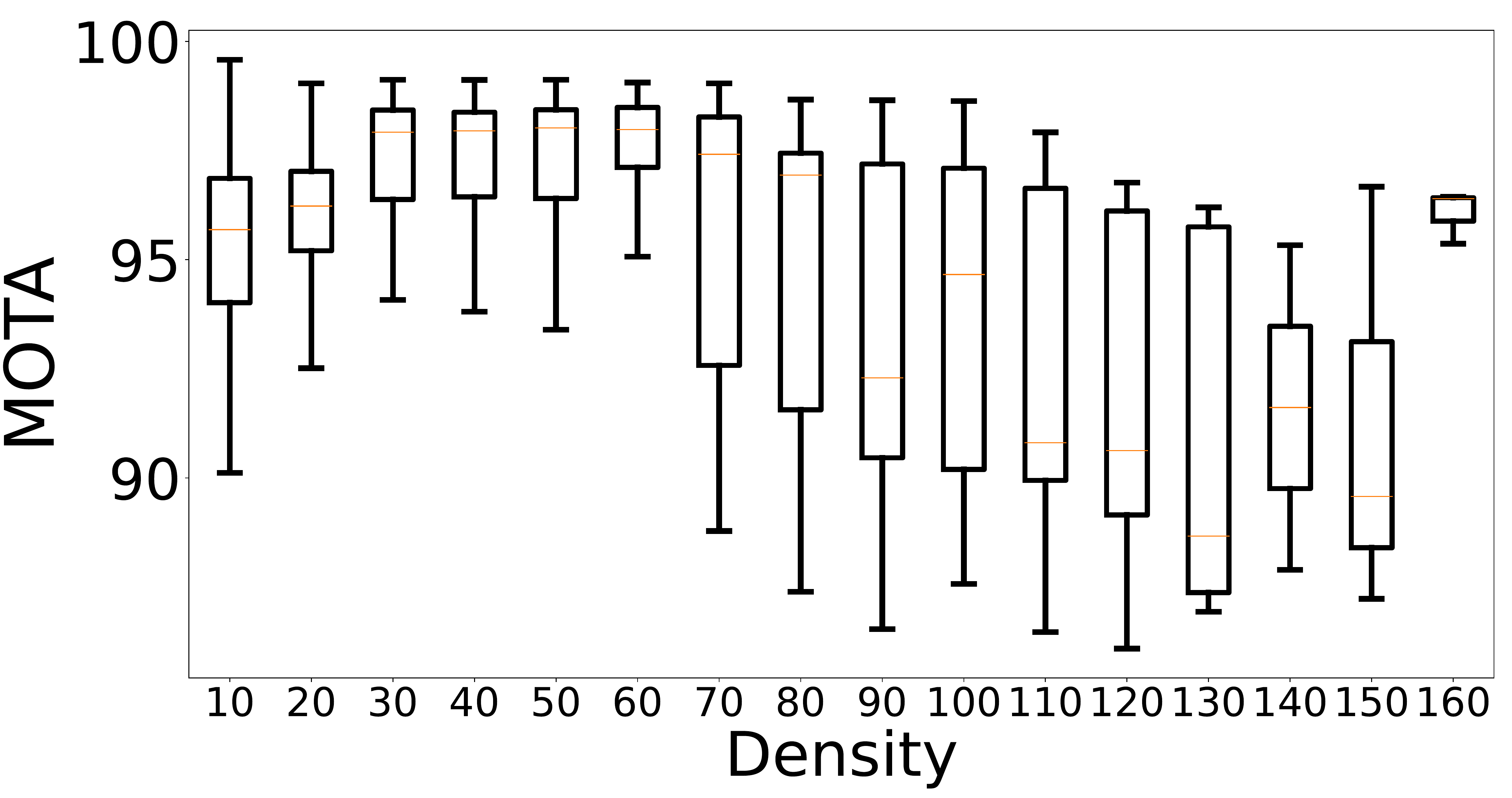}
  \caption{Varying density for MOTA on Deep-Sort.}
 \end{subfigure}
 \begin{subfigure}[b]{0.48\textwidth}
  \includegraphics[width=\textwidth]{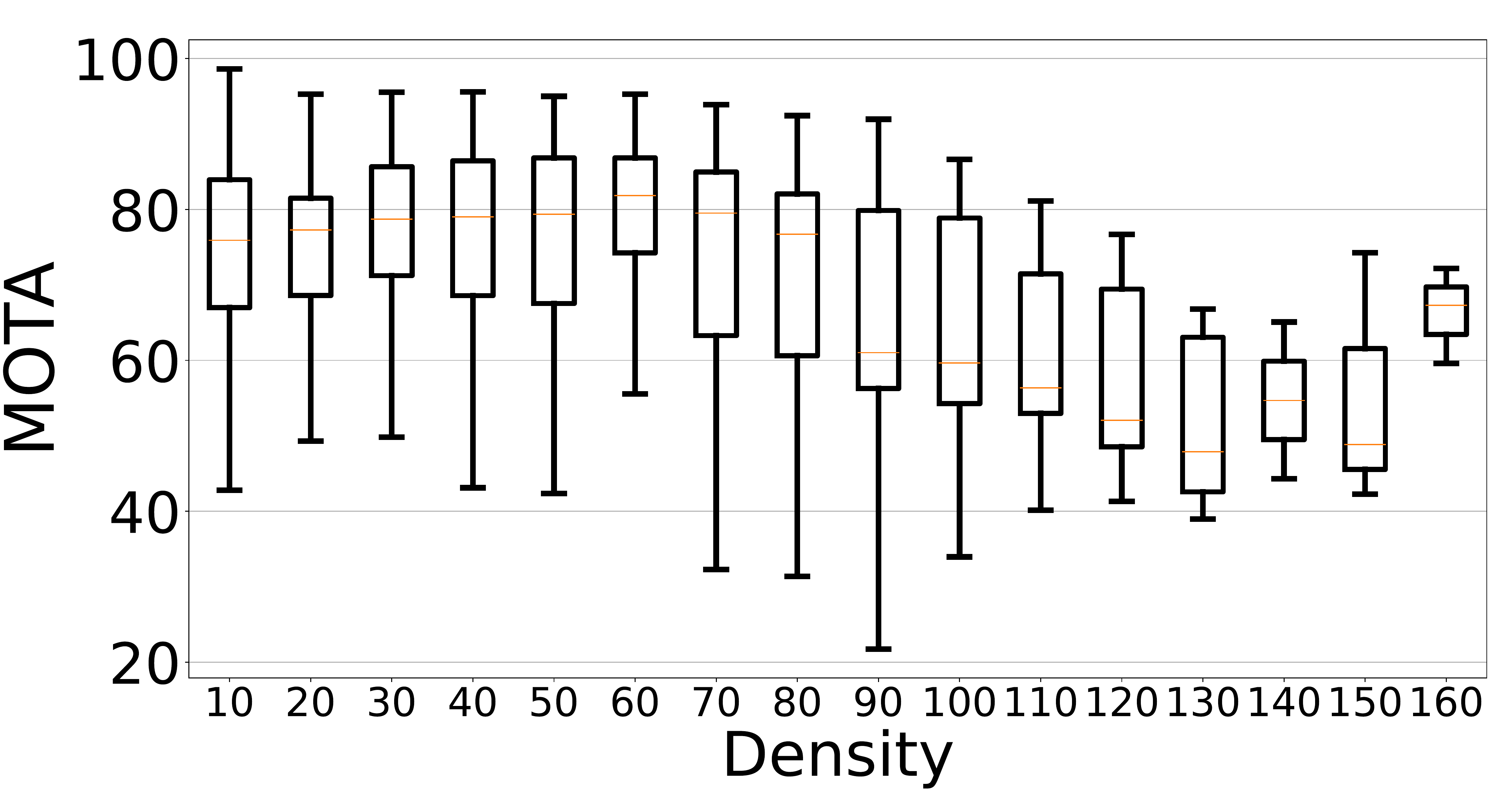}
  \caption{Varying density for MOTA on Deep-TAMA.}
 \end{subfigure}
 \caption{Results of investigated tracking methods (IOU-tracker, Deep-Sort and Deep-TAMA) on varying crowd density data (from 1 - 160 agents) on MOTA parameter.}
 \label{fig:res_density}
\end{figure*}


\section{Results and discussion}
\subsection{Collected Dataset}
CrowdSim2 was used as the main tool for the generation of many different simulated situations in 3 main places in the virtual city. All places were observed by 3 cameras directed at the same position to get the general view from different angles. All videos were recorded in the resolution of 800x600 in 25 frames per second. Each situation lasted 30 seconds - 750 pictures were recorded. The number of pedestrians varied from 1 to 160, and situations were influenced by weather conditions such as sun, rain, fog, and snow. The dataset was generated in the format of MOT Challenge. The number of generated videos, folders and frames is presented in Table \ref{tab:nums}.

\begin{figure*}[!h]
 \centering
 \begin{subfigure}[b]{0.48\textwidth}
  \includegraphics[width=\textwidth]{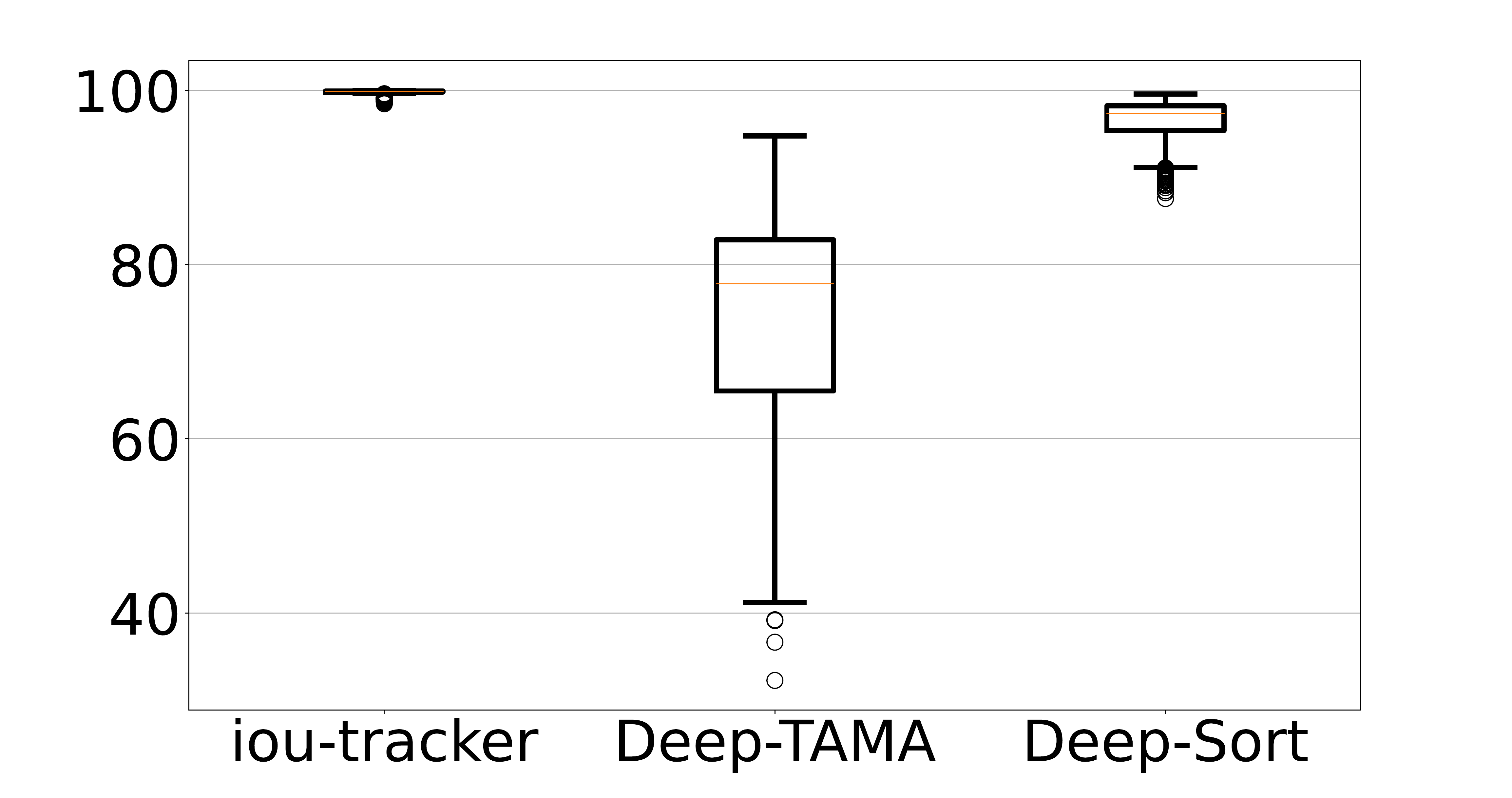}
  \caption{Fog weather condition.}
 \end{subfigure}
  \begin{subfigure}[b]{0.48\textwidth}
  \includegraphics[width=\textwidth]{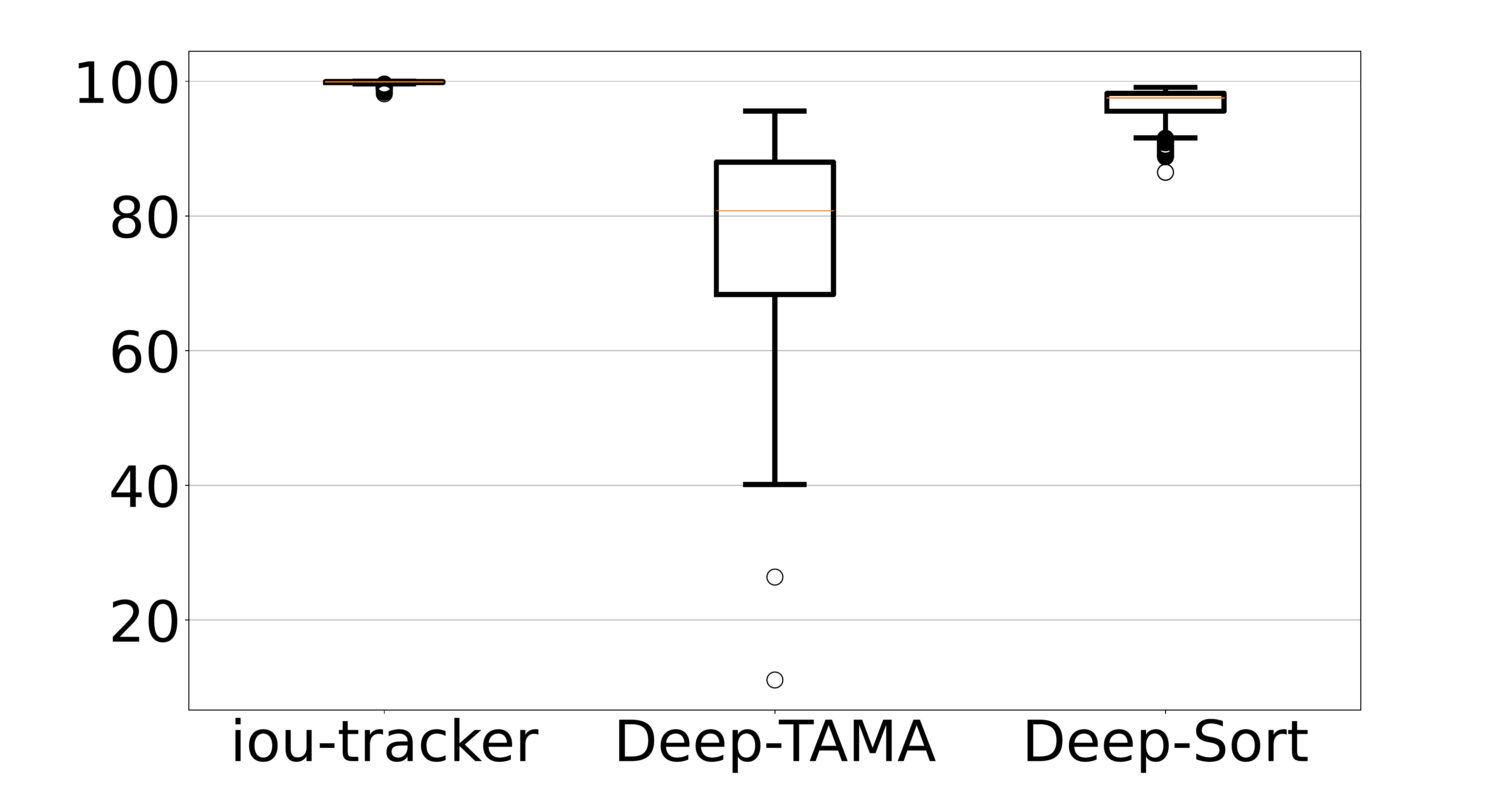}
  \caption{Rain weather condition.}
 \end{subfigure}
 \begin{subfigure}[b]{0.48\textwidth}
  \includegraphics[width=\textwidth]{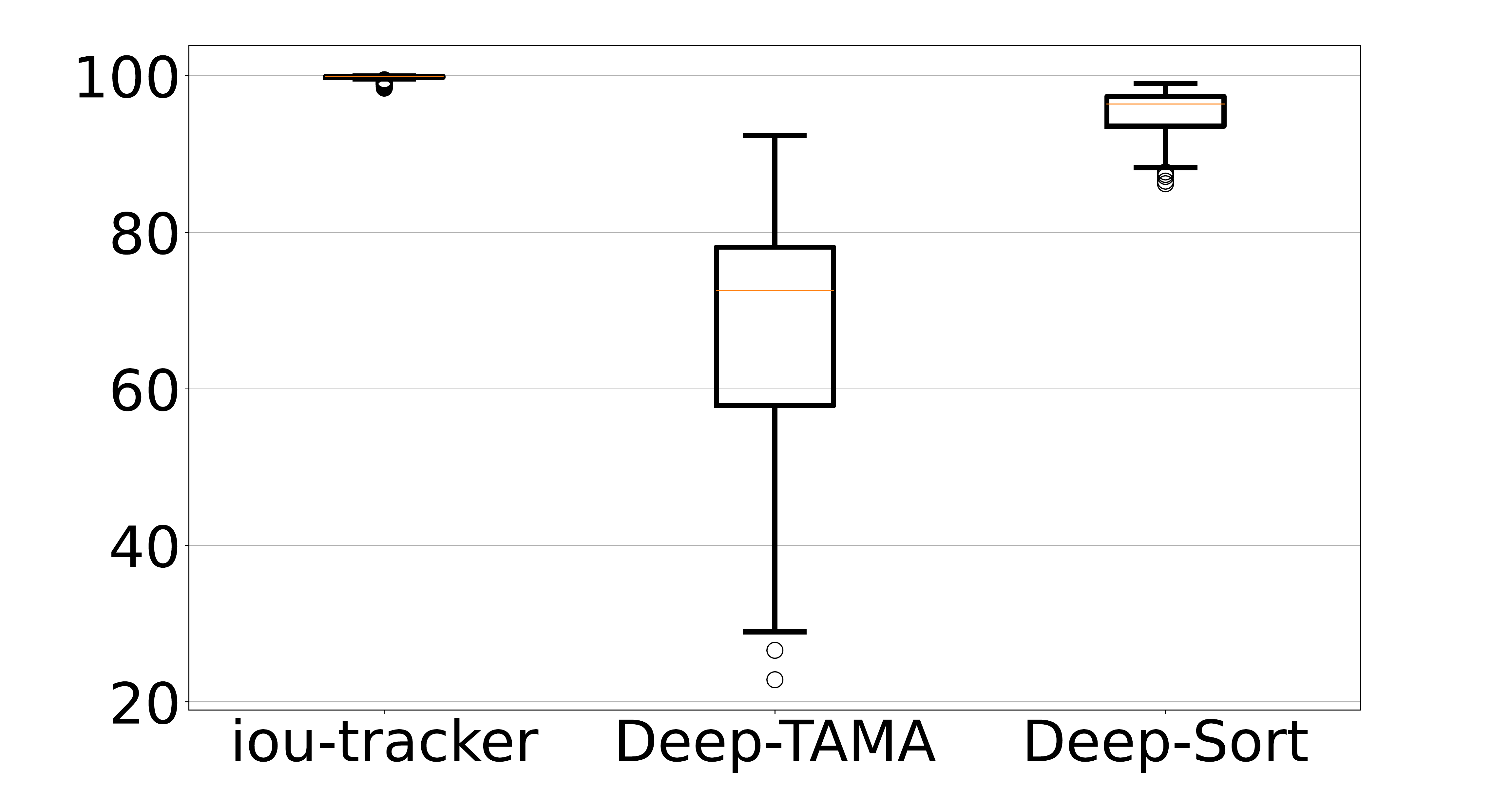}
  \caption{Snow weather condition.}
 \end{subfigure}
 \begin{subfigure}[b]{0.48\textwidth}
  \includegraphics[width=\textwidth]{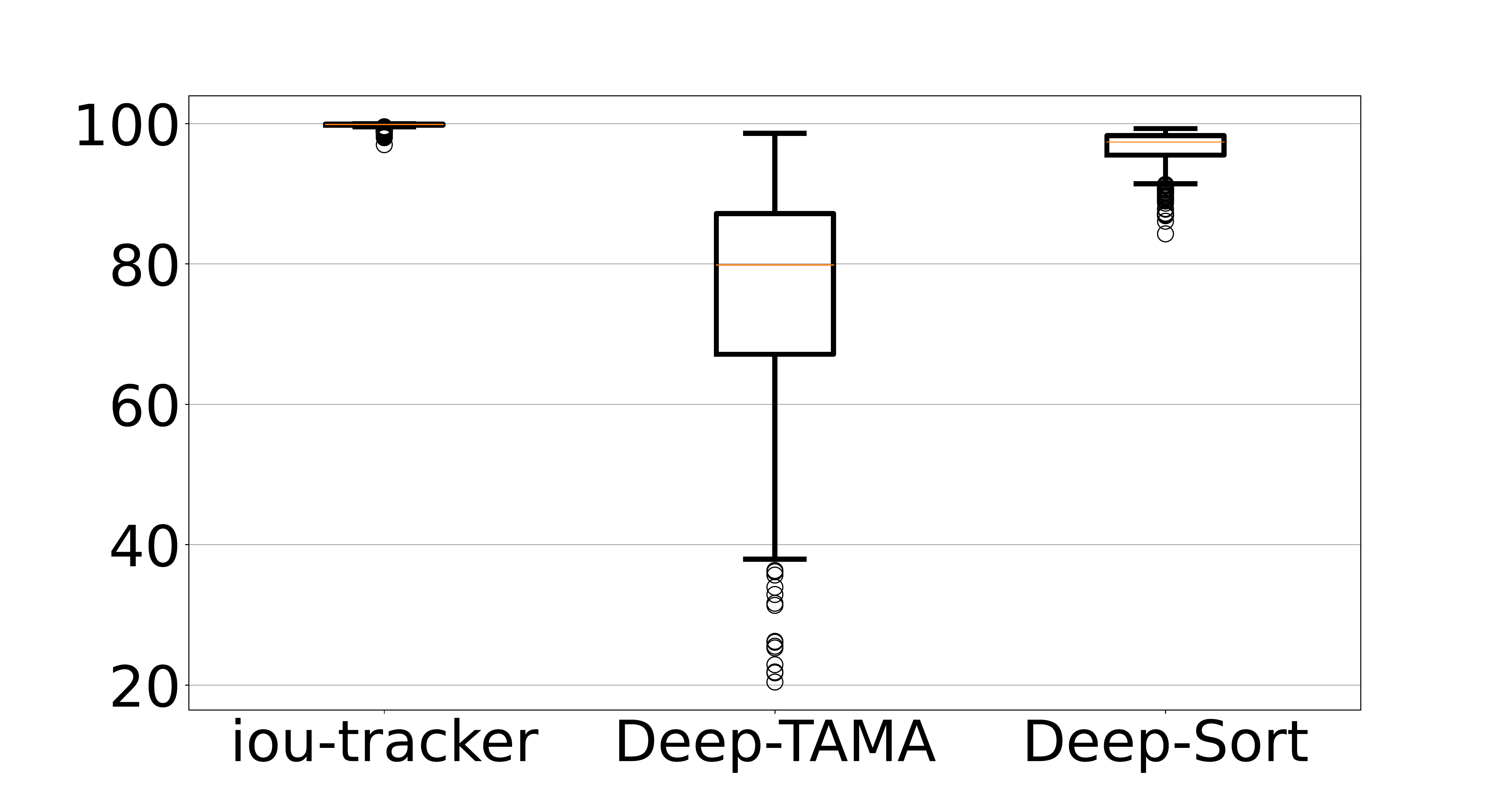}
  \caption{Sun weather condition.}
 \end{subfigure}
 \caption{The influence of different weather conditions (fog, rain, snow, and sun) on the MOTA parameter for exemplary data from CrowdSim2 presented in the form of boxplots.}
 \label{fig:res_weather}
\end{figure*}

\subsection{Results}
The obtained dataset was validated in two different scenarios: 1) by verification of the influence of the crowd density on the accuracy of tracking methods and 2) by application of 4 different weather conditions utilizing clear sunny weather (with just resulting sun reflections), rain and snow with cloudy weather influencing also background of the scene and fog day. Finally, 3 methods of tracking were used to test generated dataset: IOU-Tracker \cite{1547Bochinski2018}, Deep-Sort \cite{Wojke2018deep} and Deep-TAMA \cite{DBLP:journals/corr/abs-1907-00831}. Presented methods were chosen under two conditions - first the availability of open source code and second finite time of execution. All methods were applied in MOT Challenge ranking. The given evaluation was divided into two separate subsections - 1) validation of tracking methods in changing the number of people in simulation and 2) verification of different weather conditions. Methods were tested for the chosen parameters applied in MOT Challenge: a) MOTA - Multiple-Object Tracking Accuracy and b) IDs - ID switches.

\begin{figure*}[!h]
 \centering
  \begin{subfigure}[b]{0.48\textwidth}
  \includegraphics[width=\textwidth]{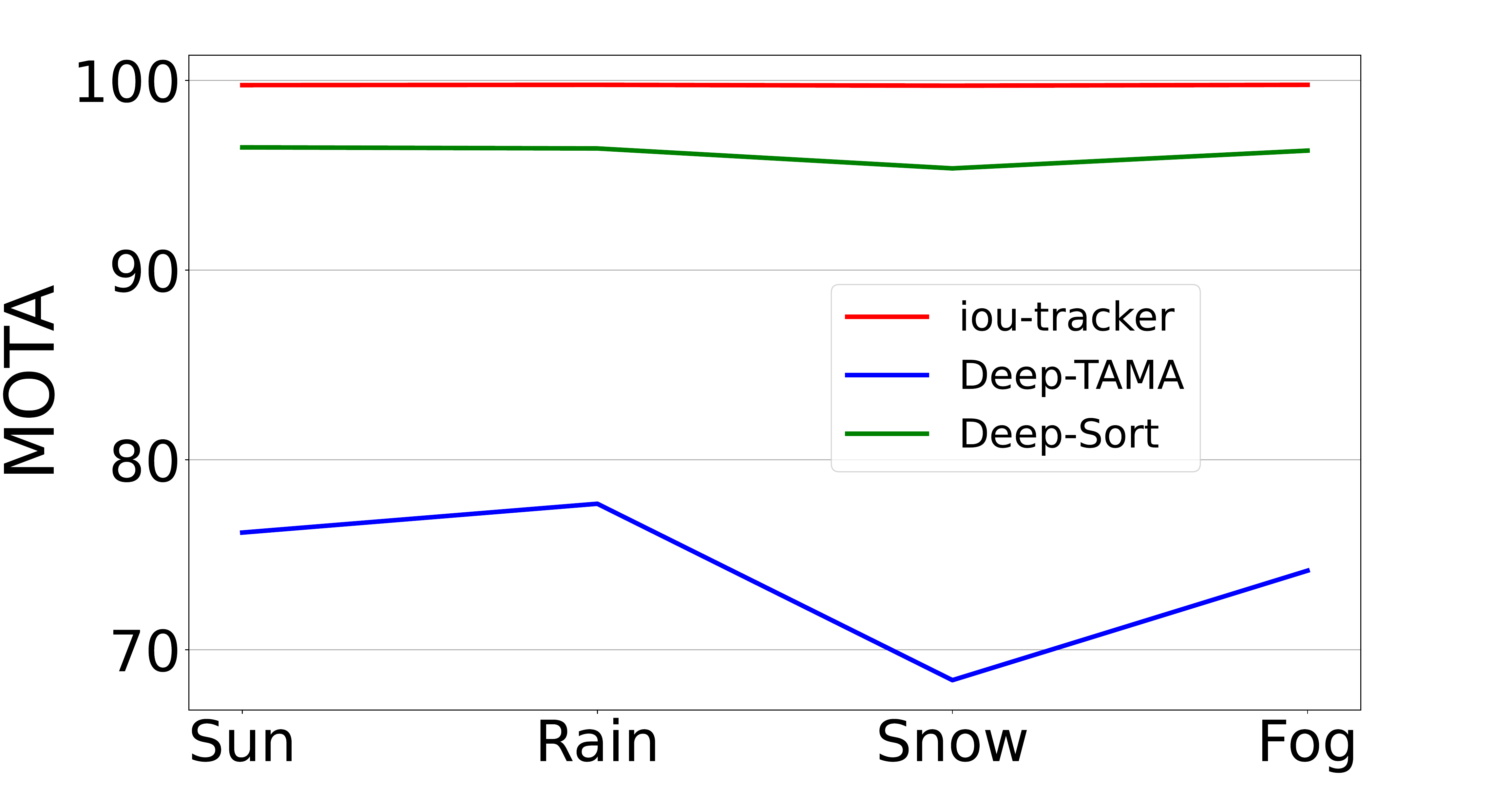}
  \caption{MOTA - Multiple-Object Tracking Accuracy.}
 \end{subfigure}
  \begin{subfigure}[b]{0.48\textwidth}
  \includegraphics[width=\textwidth]{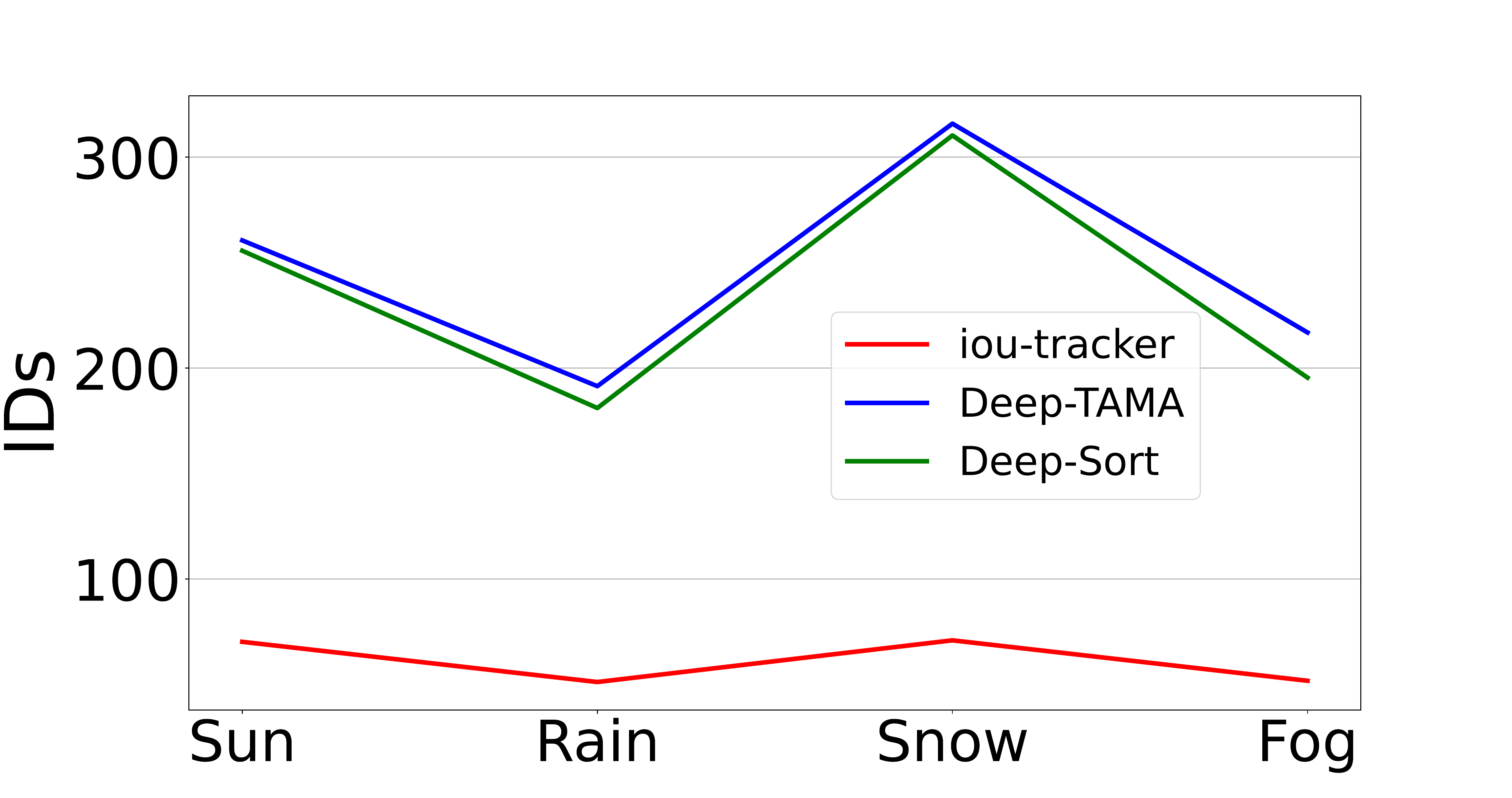}
  \caption{IDs - ID switches.}
 \end{subfigure}
 \caption{The impact of weather conditions in the validation of tracking methods for different evaluation parameters for the dataset from CrowdSim2. For MOTA the obtained results should be maximized and for IDs, the values should be lowered. }
 \label{fig:tracks}
\end{figure*}

Crowd simulation was first run on the different numbers of people, which varied from 1 to 160. Thanks to that it was possible to verify what is the influence of the number of people in tracking results. The result of the comparison is presented in Figure \ref{fig:res_density} in the form of influence on the MOTA parameter and also the distribution of results for the set of people. In the second step, data were divided into weather conditions - sun, rain, fog, and snow. Here it is also possible to distinguish differences in the results of methods. The final score was presented in a few aspects - in the form of boxplots for each weather condition on the MOTA parameter (Figure \ref{fig:res_weather}) and on different parameters in Figure \ref{fig:tracks} and Table \ref{tab:summary}.

\begin{table}
\centering
\caption{Mean and standard deviation for tracking method results concerning exemplary evaluation parameters MOTA (that should be high) and Ids (that should be low).}
\label{tab:summary}
{%
\begin{tabular}{|c|c|c|}
\hline
Method & MOTA $\uparrow$  & Ids $\downarrow$  \\ \hline
Deep-Sort  & 96.20 $\pm$ 2.64            & 362.05 $\pm$     436.04                          \\ \hline
IOU-tracker & 99.74       $\pm$ 0.36        & 13.07    $\pm$ 159.64                                \\ \hline
Deep-TAMA  & 74.58            $\pm$ 13.36      & 241.41         $\pm$ 349.84                       \\ \hline
\end{tabular}
}
\end{table}

\subsection{Discussion}
The generated data was used for two different validations of tracking methods. On one side crowd density can be the first point of analysis. For the MOTA parameter, it can be observed which method can obtain better results. In all cases, the IOU-tracker generates better results which rely mainly on the assumption that automatically all detections are given at once. Deep-Sort has slightly worse results because it takes also image context into consideration. Deep-TAMA fails in the case of simulated data due to the size of smaller detections. It has to be mentioned that the number of people is not uniformly distributed - the generation was run in the specific number of pedestrians but not always all detections were present and because of it for 160 pedestrians, not so many trails we could obtain. 

In terms of weather conditions, the order of accuracy of methods fits the crowd density analysis. The most challenging conditions are given for snow. It may result from the influence of remaining snow on the background. The same trend is visible and satisfied for different parameters from the MOTA challenge. Still in that case the best results are achieved by the IOU-tracker taking into consideration just the bounding box position. In fact to get the conclusion regarding validation both approaches can be used and still conclusion of which method can give better tracking is possible. In all cases, in terms of MOTA, it can be observed that a better method generates higher results. On the other hand for IDs, that value should be lowered. 

Crowd simulations and direct connection to a graphical engine allow the generation of very accurate detections (bounding boxes) for all visible pedestrians and on each frame. After validation performance, it could be noticed that generation of all detections has some disadvantages. The first drawback lies in the number of detections and their size - that means that also very small pedestrians can be generated and put in the ground truth. On the other hand, in practice, obtaining detection for each frame is mostly not possible. Due to evaluation processing, any changes in that protocol were not submitted but in the future generated data could be randomly disturbed in terms of the number of detections. On the other hand on generated data, some other detection methods could be applied to produce not ideal detections.

\section{Summary}
In the presented work an extension of crowd simulation CrowdSim2 was introduced with many advanced features applied for the reality enhancement of generated results. To show the practical application of generated simulated data, tracking methods were run for evaluation purposes. Algorithms were tested concerning the crowd density and weather conditions showing differences in final results and ordering accuracy of methods. The obtained results confirmed that synthetic data from CrowdSim2 can be used in the validation process for many scenarios without the need for real data. Besides tracking algorithms it can be applied for object detection, action detection, and recognition, as a part of the testing procedure, and also in the training of machine learning algorithms. In the future, generated simulated dataset can be enhanced by the used of post-processing methods to improve reality.

\section*{Acknowledgements}
This work was supported by: European Union funds awarded to Blees Sp. z o.o. under grant POIR.01.01.01-00-0952/20-00 “Development of a system for analysing vision data captured by public transport vehicles interior monitoring, aimed at detecting undesirable situations/behaviours and passenger counting (including their classification by age group) and the objects they carry”); EC H2020 project ``AI4media: a Centre of Excellence delivering next generation AI Research and Training at the service of Media, Society and Democracy'' under GA 951911; research project (RAU-6, 2020) and projects for young scientists of the Silesian University of Technology (Gliwice, Poland); research project INAROS (INtelligenza ARtificiale per il mOnitoraggio e Supporto agli anziani), Tuscany POR FSE CUP B53D21008060008. Publication supported under the Excellence Initiative - Research University program implemented at the Silesian University of Technology, year 2022.
This research was supported by the European Union from the European Social Fund in the framework of the project "Silesian University of Technology as a Center of Modern Education based on research and innovation” POWR.03.05.00- 00-Z098/17.
We are thankful for students participating in design of Crowd Simulator: P. Bartosz, S. Wróbel, M. Wola, A. Gluch and M. Matuszczyk.

\normalsize
\bibliography{references}

\end{document}